\documentclass[authoryear,preprint,review,12pt]{elsarticle}

\usepackage{doi}

\usepackage{amssymb}
\usepackage{amsmath}
\usepackage{float}
\usepackage{lineno}
\usepackage{xcolor}
\usepackage{natbib}
\usepackage[flushleft]{threeparttable}
\usepackage{longtable}

\journal{TBD}

\begin{document}

\begin{frontmatter}



\title{Self-supervised and Multi-fidelity Learning for Extended Predictive Soil Spectroscopy} 


\author[1]{Luning Sun\corref{corxh}}
\ead{sun42@llnl.gov}
\author[2]{José L. Safanelli}
\author[2]{Jonathan Sanderman}
\author[3]{Katerina Georgiou}
\author[4]{Colby Brungard}
\author[4]{Kanchan Grover}
\author[5]{Bryan G. Hopkins}
\author[1]{Shusen Liu}
\author[1]{Timo Bremer}

\cortext[corxh]{Corresponding author}
\affiliation[1]{organization={Lawrence Livermore National Laboratory},
            city={Livermore},
            state={CA},
            country={USA}}

\affiliation[2]{organization={Woodwell Climate Research Center},
            city={Falmouth},
            state={MA},
            country={USA}}

\affiliation[3]{organization={Oregon State University},
            city={Corvallis},
            state={OR},
            country={USA}}

\affiliation[4]{organization={New Mexico State University},
            city={Las Cruces},
            state={NM},
            country={USA}}

\affiliation[5]{organization={Soil Science Society of America — North American Proficiency Testing Program, and Brigham Young University},
            city={Provo},
            state={UT},
            country={USA}}

\begin{abstract}
We propose a self-supervised machine learning (SSML) framework for multi-fidelity learning and extended predictive soil spectroscopy based on latent space embeddings. A self-supervised representation was pretrained with the large USDA NSSC-KSSL mid-infrared (MIR) spectral library and the Variational Autoencoder algorithm to obtain a compressed latent space (32 features) for generating spectral embeddings. At this stage, only unlabeled spectral data were used, allowing us to leverage the full spectral database and the availability of scan repeats for augmented training (n scans = 334,665). We also leveraged and froze the trained MIR decoder for a spectrum conversion task by plugging it into a near-infrared (NIR) encoder to learn the mapping between NIR and MIR spectra in an attempt to leverage the predictive capabilities contained in the large MIR library with a low cost portable NIR scanner. This was achieved by using a smaller subset of the KSSL library, which contains paired NIR and MIR spectra (n = 2,106). Downstream machine learning models were then trained to learn the mapping between original spectra, predicted spectra, and latent space embeddings for nine soil properties: total carbon (TC), total nitrogen (TN), inorganic carbon (IC), estimated organic carbon (EOC), soil pH (pH), cation exchange capacity (CEC), clay, silt, and sand. The performance of soil property prediction was evaluated independently of the KSSL training data using a gold-standard test set derived from the North American Proficiency Test (NAPT) Program soil archive, along with regression goodness-of-fit metrics. Compared to baseline models, the proposed SSML and its embeddings yielded similar or better accuracy in all soil properties prediction tasks. Predictions derived from the spectrum conversion (NIR to MIR) task did not match the performance of the original MIR spectra but were similar or superior to predictive performance of NIR-only models, suggesting the unified spectral latent space can effectively leverage the larger and more diverse MIR dataset for prediction of soil properties not well represented in current NIR libraries.


\end{abstract}




\end{frontmatter}
 


\section{Introduction}
\label{sec:intro}

Measuring and monitoring soils at scale is costly \citep{Mercer2024}. The demand for high-quality soil information is surging with the increased recognition of the soil's role in climate regulation and adaptation, sustainable agricultural production, and its importance in several other ecosystem services \citep{Amundson2015}. Emerging technologies for generating basic soil data in response to this growing demand have been maturing in academia and are now being tested in commercial setups \citep{Poppiel2022}. Among the range of potential measurement technologies, diffuse reflectance spectroscopy (DRS) is one of the most promising \citep{ViscarraRossel2022, Ng2022}, as the spectral acquisition of a soil sample is fast (often takes less than a minute), requires minimum sample preparation, does not generate chemical residues, and is sensitive to some soil properties variations \citep{shepherd2022global, Safanelli2025_OSSL, Sanderman2023}. This happens because the resulting DRS spectra are characterized by a combination of interactions between light and organic, mineral, and sample bulk constituents, which can be explored using machine learning when analytical reference data are available. Still, challenges related to handling spectral data more effectively, addressing variable fidelity between spectral representations and relevant soil properties, and increasing the reliability of generated predictions, all remain as research opportunities to unlock this technology as a viable soil measurement tool.

DRS spectra measure the amount of light reflected from the surface of a soil sample in relation to a reference material. This interaction can be measured across the visible (Vis, 400-700 nm), near (NIR, 700-2500 nm), and mid-infrared electromagnetic region (MIR, 2500-25000 nm, i.e., 4000-400 cm$^{-1}$), depending on the instrument characteristics and study goals. The measured signal is represented by its intensity as a function of evenly spaced contiguous channels, measured across the electromagnetic range of interest (e.g., VisNIR, NIR, or MIR). Therefore, soil spectral data are inherently high-dimensional (hundreds of wavelengths scanned contiguously), multicollinear (nearby and distant wavelengths are strongly correlated), hence redundant \citep{Pasquini2003}. Multicollinearity of spectral data can be partially addressed through preprocessing techniques, such as employing the Standard Normal Variate (SNV) method \citep{Barnes1989}, which centers and scales each spectrum to its origin and unit across the wavelength range, respectively. Still, remaining redundancy and the high-dimensionality issues often require specific strategies to calibrate more consistent prediction models. Partial Least Squares Regression (PLSR) and its variants are examples of multivariable and/or multivariate algorithms that are capable of decomposing the high dimensionality and redundancy of spectral data via orthogonalization while simultaneously maximizing the variance of a single or a set of target variables, which are later combined with multiple linear regression \citep{Geladi1986}. PLSR is considered the standard fitting algorithm in chemometrics or machine learning with spectral data, and is often used as a baseline. Similarly, other linear orthogonalization algorithms like principal component analysis (PCA) have also been combined with statistical or machine learning algorithms in predictive soil spectroscopy to control these issues, which often benefits model performance \citep{Vestergaard2021, Barra2021, Safanelli2025_OSSL}.

With the advent of deep learning, which offers greater flexibility in designing complex architectures and loss functions for exploring linear and non-linear relationships of data \citep{Minasny2024}, several architectures have been tested in soil spectroscopy. This includes simpler models with multilayer perceptrons \citep{Wijewardane2016, Wijewardane2018} to more sophisticated architectures like convolutional and recurrent neural networks \citep{Ng2019, Padarian2019, Tsakiridis2020, Yang2020, Omondiagbe2023}, as well as novel algorithms based on autoencoders and transformers that are forming the basis of foundation models in several fields, including soil spectroscopy \citep{Ke2024, Guo2025}. A paradigm that has recently emerged is self-supervised learning, where a model is trained on unlabeled data \citep{pmlr-v27-baldi12a}. This is especially interesting for soil spectroscopy, as an autoencoder can effectively compress the spectra into a latent space (i.e., a lower-dimensional representation) to produce discriminative features (embeddings) while eliminating redundancy and reducing the high dimensionality of the spectral data by orders of magnitude. \cite{Tsimpouris2021} explored this type of architecture and effectively demonstrated the enhancement of prediction accuracy based on soil spectral data.

In this paper, we leverage self-supervised learning on a large MIR soil spectral library to learn a latent feature space of 32 dimensions. We extend our study from previous work with an additional hypothesis: the latent space represents a manifold of the fundamental infrared spectral region, enabling interoperability between different fidelities, i.e., NIR and MIR (Fig.~\ref{fig:latent}). This hypothesis was explored because the NIR range contains broad features, overtones, and combinations of overtones from the fundamental MIR range at higher frequencies \citep{Pasquini2003, Weyer1985}. Hence, NIR spectral features are highly overlapping and less sensitive to soil variations (low-fidelity), such that making a bridge between the NIR and MIR could extend or enhance the predictive capacity of a low-fidelity scanner with access to larger MIR datasets. This is especially compelling for scaling soil spectroscopy as a measuring and monitoring tool because low-fidelity NIR scanners are an order of magnitude cheaper than bench-top MIR instruments (high-fidelity), and recent hardware advances have enabled the miniaturization of more sensitive and stable sensors based on Fourier Transform (FTIR) technology \citep{Pasquini2018}.


 \begin{figure}[H]
   \centering
   \includegraphics[width=0.71\textwidth]{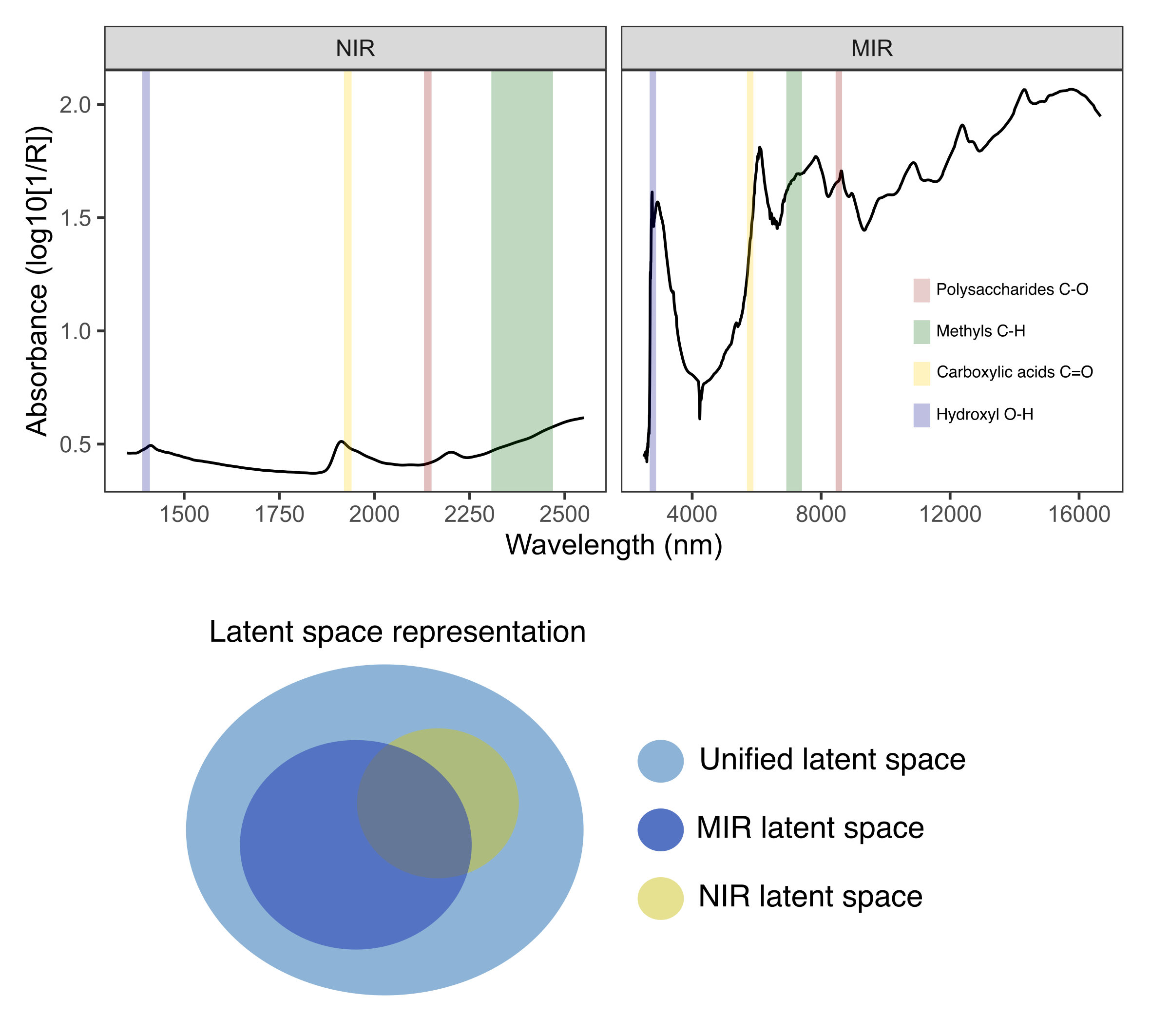}
   \caption{Illustrative example of functional organic groups and their respective spectral bands at the fundamental mid-infrared (MIR) range and their overtones in the near-infrared (NIR) range for a random soil sample (top): fundamental band for Polysaccharides C-O group is positioned around 1170 cm$^{-1}$ (8547 nm) with its respective fourth overtone represented around 2137 nm; fundamental band for Methyls C-H group is positioned between 1445-1350 cm$^{-1}$ (6920-7407 nm) with its respective third overtone represented between 2307-2469 nm; fundamental band for Carboxylic acids C=O group is positioned around 1725 cm$^{-1}$ (5797 nm) with its respective third overtone represented around 1930 nm; fundamental band for Hydroxyls H-O group is positioned around 3600 cm$^{-1}$ (2778 nm) with its respective second overtone represented around 1400 nm \citep{Weyer1985}. This physical relationship of fundamental vibrations and overtones across the infrared range forms the basis of a unified latent space hypothesis (bottom).}
   \label{fig:latent}
 \end{figure}

This study aimed at i) pre-training an autoencoder model for the fundamental MIR spectra as a self-supervised learning task with scan repeats (augmentation); ii) learning a multi-fidelity representation between NIR spectra (low-fidelity) with the pre-trained fundamental MIR latent space (high-fidelity) to make a bridge between both; and iii) evaluating the predictive performance of latent space embeddings and converted spectra with downstream regression of soil properties.
\section{Material and Methods}
\label{sec:method}
\subsection{Training and validation data}
The majority of spectral data used in this study was sourced from the USDA-NRCS-NSSC-KSSL database (hereafter referred to as KSSL). This database consists of tens of thousands of soil samples collected throughout the United States and other territories over the past decades that were consistently processed and analyzed following the Kellogg Soil Survey Laboratory (KSSL) Methods Manual \citep{KSSL}. Spectral data, specifically from the mid-infrared range (MIR), were collected with a Fourier-Transform Infrared spectrometer (FTIR, Vertex 70, Bruker Optics) in diffuse reflectance mode (DRIFT). The instrument is composed of a mercury cadmium telluride (MCT) detector cooled by liquid nitrogen, a potassium bromide (KBr) beam splitter, and gold mirrors. Roughened aluminum is used as a reference material for calculating reflectance. Soil samples were air-dried, ground, sieved to $\textless$2 mm, and further ground to 180 $\mu$m and then loaded into the plate wells in four repeats (subsamples or aliquots) for each soil sample ID. Soil samples are pressed on a 96-well spot plate and scanned with the Bruker HTS-XT diffuse reflectance accessory. The scanning compartment is not purged. The resulting spectra are given with a spectral resolution of 4 cm\textsuperscript{-1}, with the spectra of each well being derived from 32 co-added instantaneous scans computed by the instrument \citep{Wijewardane2018}. It is important to note that we have leveraged the subsample scan repeats during the self-supervising learning stage. In contrast, the average scan was employed in the soil properties regression tasks. In this stage, 83,971 distinct soil samples with a total of 334,665 scan repeats were used for the self-supervised learning. A random 80/20 split was used during the learning stage for model optimization and verification.

For the multi-fidelity learning, a subset of the KSSL MIR library was scanned with the ProxiScout\textsuperscript{TM} Portable NIR (1350-2550 nm) developed by Buchi (formerly known as Neospectra\textsuperscript{TM} from Si-Ware). Each scanner is equipped with Fourier-Transform NIR technology (FT-NIR), based on the semiconductor Micro-Electromechanical Systems (MEMS) manufacturing technique, which promises accuracy and consistency between devices. The library comprises 2,106 distinct mineral soil samples scanned across several portable scanners (built in 2022). However, only 1976 soil samples contain both the NIR and MIR paired spectra (hereafter referred to as paired). The soil samples from this dataset were selected to represent the diversity of mineral soils found in the United States and some countries in Africa. For the spectral measurement, all samples were scanned dry after being sieved through a 2 mm sieve. Approximately 20 g of sample was added to a plastic weighing boat, where the scanner was placed to make direct contact with the soil surface. The scanner was gently moved across the sample surface and six replicate scans were taken to control for surface variability. A white reference cap was used as a reference material for instrument calibration, allowing the computation of the diffuse reflectance of soil samples. In this dataset, some soil spectra contained both scan and instrument repeats for NIR measurements; however, this distinction was disregarded in our analysis, and the actual variations were averaged across all available repeats to pair each NIR spectrum to its MIR counterpart. For more information, \cite{Partida2025} provides a comprehensive description of the NIR dataset. Lastly, a random split of 20\% was retained before training for optimizing and validating the conversion model. The performance metrics from the validation samples were also provided and compared to those from an independent test to check for potential shifts in spectral feature representation.

Both the latent space embeddings and the multi-fidelity converted spectra were tested in downstream regression tasks to understand their applications for predicting soil properties. The output features from the previous tasks were combined with reference analytical data available in the KSSL database. This database has several analytes linked to soil health and general soil characteristics, ranging from common physical and chemical soil properties necessary for informed crop production to biological and mineralogical characterization \citep{Sanderman2020, Ng2022}. Given that several studies have already demonstrated the potential of predictive soil spectroscopy in estimating a range of soil properties, e.g., \citep{nocita2015soil, SorianoDisla2013, Safanelli2025_OSSL}, we decided to focus on nine soil properties that were recently evaluated with additional effectiveness criteria beyond the usual goodness-of-fit regression metrics \citep{Grover2025}, i.e., comparing the predictions with measurement control and warning ranges defined by a proficiency testing program. The set of soil properties includes: total carbon (TC), total nitrogen (TN), inorganic carbon (IC), estimated organic carbon (EOC), soil pH, cation exchange capacity (CEC), clay, silt, and sand. The analytical procedures of these soil properties followed the KSSL Methods Manual, with TC and TN being determined from dry combustion and given as weight percentage; IC is defined as CaCO3 equivalent as weight percentage; EOC is defined by the subtraction of IC from TC ($EOC=TC-0.12IC$); pH is determined from a 1:1 soil:water solution; CEC was measured by cation displacement, given as cmol\textsubscript{c} kg\textsuperscript{-1}; and particle size distribution (clay, silt, and sand percent) determined by the pipette method and presented as weight percentage \citep{KSSL}.

For an independent evaluation of the proposed modeling strategies (not used during the training stage), a golden test set compiled by \cite{Grover2025} and sourced from the North America Proficiency Testing (hereafter referred to as NAPT) program's soil archive was used in this study. The median measurement value of numerous laboratories was used to approximate the actual value for each analyte (ranging from 5 to 150 labs per quarterly test over the years). The NAPT soil data encompass several soil properties and methods; therefore, only those matching the KSSL analytical methods were used in this evaluation. Spectral data were collected in slightly different conditions but are still compatible with the previous datasets: MIR spectra were acquired using a Bruker Invenio-R FTIR spectrometer with an HTS-XT accessory, also equipped with an MCT detector cooled by liquid nitrogen. Air-dried and sieved (\textless2 mm) soil samples were finely ground to a particle size of approximately 180 $\mu$m and loaded into a 96-well microplate. Four subsamples/aliquots were drawn from each soil sample and loaded into separate wells on the microplate. The spectrum of each subsample was obtained by averaging 32 co-scans, maintaining a spectral resolution of 4 cm\textsuperscript{-1}. A gold standard was used as reference material for calculating the diffuse reflectance. For the NIR spectra, a newer ProxiScoutTM Portable NIR (1350-2550 nm) (formerly known as NeospectraTM) was used with the same standard operating procedures as used for the NIR dataset generation, which includes the scanning of dry and sieved (<2 mm) soil samples in the lab. NIR spectra, however, were trimmed to the 1400-2500 nm range in all NIR datasets to remove some visual spectral variations caused by the change of instrument configuration.

Summary statistics of the nine soil properties across the three different datasets (KSSL, Paired, and NAPT) are provided in \ref{app_soil}.

\subsection{Pre-training and self-supervised learning}
Based on the hypothesis of a unified latent space, a lower-dimensional manifold learned from a high-dimensional dataset can be mathematically defined as follows:
\begin{equation}
\mathbf{z} = \mathcal{F}(\mathbf{X})
\end{equation}
where $z \in \mathbb{R}^m$ and $X \in \mathbb{R}^d$, $m < d$. $\mathcal{F}$ is a functional mapping that compresses the data and can be a linear mapping, such as principal component analysis (PCA), or a nonlinear function, like deep neural networks (DNN). It usually consists of stacks of NN layers, e.g., Multilayer Perceptron (MLP), Convolutional Layers (CNN). In the self-supervised learning for $\text{MIR}$ spectrum, we adopt a encoder-decoder-based architecture, given as:
\begin{align}
\mathbf{z}_{\text{MIR}} &= \mathcal{F}_1^{ENC}(\mathbf{X}_\text{MIR})\\
\mathbf{\hat{X}}_\text{MIR} &= \mathcal{F}_1^{DEC}(\mathbf{z})
\end{align}
We get the low-dimensional hidden space by learning an identity mapping $\mathbb{R}^{d}\mapsto\mathbb{R}^{d}$ between reconstructed data $\mathbf{\hat{X}}$ and data $\mathbf{X}$. \\

\subsection{Multi-fidelity learning}
Mapping the NIR and MIR to a unified latent space is attractive for two reasons. Firstly, the latent space could capture the shared features across the multi-fidelity dataset. Secondly, scanning new soils with NIR is less expensive and faster than scanning with MIR. This would allow soil spectroscopy to scale as a measuring and monitoring tool for natural resource assessment and soil health applications. In this work, we trained a separate encoder for NIR data to map the NIR to the pretrained latent space from MIR data by: 
\begin{equation}
\mathbf{z}_{\text{NIR}} = \mathcal{F}^{ENC}_2(\mathbf{X}_{\text{NIR}})
\end{equation}
We note that the loss function is defined in the original MIR space since the distance metric in the latent space does not have a clear meaning. Specifically, the loss function is defined as:
\begin{equation}
\mathcal{L}_1=||\hat{\mathbf{X}}_{\text{NIR}-\text{MIR}}-\mathbf{X}_{\text{MIR}}||_{L_2}
\end{equation}
where $\hat{\mathbf{X}}_{\text{NIR}-\text{MIR}}=\mathcal{F}_1^{DEC}({\mathbf{z}}_{\text{NIR}})$. The training process minimizes the scalar loss $\mathcal{L}$ while implicitly minimizing the discrepancy between $\mathbf{z}_\text{NIR}$ and $\mathbf{z}_\text{MIR}$. To stabilize the training process, we add a regularization loss accounting for the reconstruction error for the NIR spectrum, defined as:
\begin{equation}
\mathcal{L}_r = \alpha_t||\hat{\mathbf{X}}_{\text{NIR}}-\mathbf{X}_{\text{NIR}}||_{L_2}
\end{equation}
where $\alpha_t$ linear decreases with epochs $t$ from 1 to 0, and $\hat{\mathbf{X}}_{\text{NIR}}=\mathcal{F}_{3}^{DEC}$. It is worth noting that the $\mathcal{F}_{3}^{DEC}$ was not used for multi-fidelity spectra conversion, but serves as a regularizer to guide the training process. \\

\subsection{Predictive modeling}
For predictive purposes, a reduced latent space may be an optimal solution to control redundant, collinear, and highly dimensional spectral features and help calibrate more robust predictive models. Therefore, we construct different mappings based on the latent representation, given by:
\begin{equation}
\mathbf{Y} = \mathcal{F}_3(\mathbf{z})
\end{equation}
where $\mathbf{Y} \in \mathbb{R}^{n}$ represents soil properties. With this definition, we could finally predicting soil properties from $\text{NIR}$ to $\text{MIR}$, by evaluating $\hat{\mathbf{Y}} = \mathcal{F}_3[\mathcal{F}_2^{\text{ENC}}(\mathbf{X}_{NIR})]$. The mapping functions linking spectral data or latent space embeddings to soil properties were the PLSR \citep{Geladi1986}, used as a baseline algorithm in predictive soil spectroscopy \citep{SorianoDisla2013, Barra2021}, and multilayer perceptrons (MLP) \citep{Wijewardane2018}.

\begin{figure}[htp!]
   \centering
   \includegraphics[width=1\textwidth]{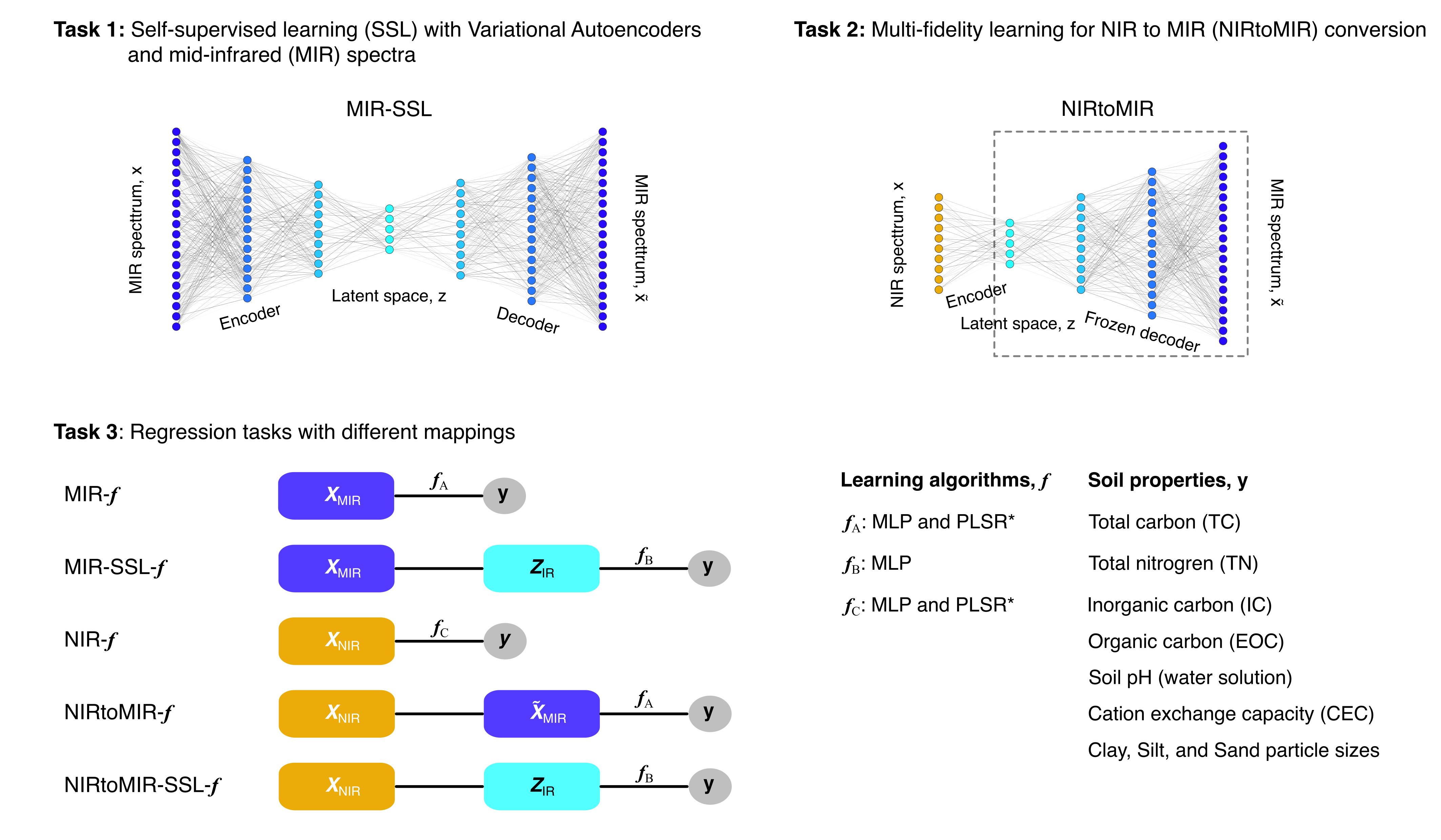}
   \caption{Learning strategies employed in this study. Firstly, the self-supervised learning is leveraged to pre-train and learn a latent space representation from mid-infrared (MIR) spectra with variational autoencoders (VAE) and scan repeats. Then, the learned MIR decoder is frozen and aligned with a near-infrared (NIR) encoder to enable the learning and mapping between the two fidelities (NIRtoMIR). Lastly, nine relevant soil properties were mapped from original spectral data, learned latent representations, and multi-fidelity outputs for predictive purposes using either partial least squares regression (PLSR) or multilayer perceptron (MLP). Please note: stacked neural networks are illustrative examples.}
   \label{fig:method}
 \end{figure}

In summary, our framework is trained in three stages (Fig.~\ref{fig:method}). In stage one, a self-supervised learning model is trained on the KSSL MIR dataset to obtain the latent space representations and the weights for the encoder and decoder mapping. In stage two, we fixed the decoder from the self-supervised learning model and trained a new encoder to map from the NIR to MIR spectrum. This setting is used to align the latent representation of both spectra. To facilitate the training process in stage two, a composite loss function was employed, incorporating both NIR reconstruction error and NIR to MIR mapping error, with an adaptive weighting coefficient. In the current work, we used a linearly decreasing weighting schedule to achieve a smooth transition. Lastly, in the third stage, soil property values were mapped from latent space embeddings. To evaluate the proposed strategies, we employed various baselines, including direct mapping from either spectral fidelities (MIR and NIR) and embeddings from MIR, as well as learning algorithms (PLSR and MLP).

\subsection{Data preprocessing and statistical analysis}

For both the training and independent test sets, some soil samples contained incomplete values for specific analytes. Therefore, we select only valid indices across all nine soil properties to ensure data completeness during the training and independent evaluation. This resulted in 51,508 observations for the MIR dataset, 2,016 observations for the paired dataset (NIR and MIR), and 206 observations for the independent test set with both NIR and MIR spectra. It is worth noting that the smaller NIR dataset was used for both NIR-to-MIR conversion and baseline model development, whereas the other mappings could leverage the higher sample size of the MIR database.

We also applied knowledge-informed transformation specifically for handling the compositional nature of particle size distribution (clay, silt, and sand) using isometric log ratio (ILR) transformation, which maps a $D$ dimension vector to a $D-1$ dimension real vectors, defined as:
\begin{equation}
\label{eq:ilr}
\text{ILR}(\mathbf{x})=[\langle \mathbf{x}, e_1\rangle, \langle \mathbf{x}, e_2\rangle ..., \langle \mathbf{x}, e_{D-1}\rangle]
\end{equation}
where $[e_1, e_2...,e_{D-1}]$ is an orthogonal basis in the simplex. The inverse transform is defined as:
\begin{equation}
\label{eq:ilr}
\text{ILR}^{-1}(\mathbf{y}) = \bigoplus_{i=1}^{D-1} y_i \odot \mathbf{e}_i
\end{equation}
where $\mathbf{y} = (y_1, y_2, \ldots, y_{D-1}) \in \mathbb{R}^{D-1} \text{ are the ILR coordinates}$. And $\mathbf{e}_i \in S^D \text{ are orthonormal basis vectors in the Aitchison simplex.} D$ is the number of compositional parts. $\odot$ denotes the power operation and $\bigoplus$ denotes the perturbation operation (compositional sum). For all the other soil properties (except pH and particle size fractions), we natural log-transformed (with offset 1) to control distribution right-skewness. Evaluation was made after back-transforming all soil properties to their original scale, including ILR.

For spectral preprocessing, we ensured that all spectra were presented in pseudo-absorbance units using the formula $\mathbf{A} = \text{log}_{10}(\frac{1}{\mathbf{R}})$, where $\mathbf{R}$ is the spectra given in reflectance percent fraction. We removed offset and scattering effects in both NIR and MIR spectra using the standard normal variate (SNV) method \citep{Barnes1989}, thereby centering every individual spectrum and its wavelengths at the origin and scaling them to unity.

To evaluate the predictive performance of soil properties, we calculated multiple model accuracy, precision, and consistency metrics. The average model error was estimated using the root mean squared error (RMSE, Eq.~\ref{eq:rmse}), while model bias was determined by the mean error (ME, Eq.~\ref{eq:bias}). Model accuracy was calculated with R\textsuperscript{2} (Eq.~\ref{eq:rsq}) and Lin's concordance correlation coefficient (CCC, Eq.~\ref{eq:ccc}), while model prediction consistency was determined by the ratio of performance to interquartile range (RPIQ, Eq.~\ref{eq:rpiq}). All metrics for all modeling combinations are reported in \ref{app_performance}, whereas only CCC and RPIQ are utilized for model comparison and discussion in the main text.

\begin{equation}
\label{eq:rmse}
\text{RMSE} = \sqrt{\frac{1}{N}||\mathbf{y}-\mathbf{\hat{y}}||^2}
\end{equation}

\begin{equation}
\label{eq:bias}
\text{ME} = -\frac{||\mathbf{y}-\mathbf{\hat{y}}||}{N}
\end{equation}

\begin{equation}
\label{eq:rsq}
\text{R}^2 = 1 - \frac{||\mathbf{y}-\mathbf{\hat{y}}||^2}{||\mathbf{y}-\bar{y}\textbf{1}||^2}
\end{equation}

\begin{equation}
\label{eq:ccc}
\text{CCC} = \frac{2 (\mathbf{y} - \bar{y}\mathbf{1})^T (\mathbf{\hat{y}} - \bar{\hat{y}}\mathbf{1})}{\| \mathbf{y} - \bar{y}\mathbf{1} \|^2 + \| \mathbf{\hat{y}} - \bar{\hat{y}}\mathbf{1} \|^2 + N(\bar{y} - \bar{\hat{y}})^2}
\end{equation}

\begin{equation}
\label{eq:rpiq}
\text{RPIQ} = \frac{Q_3(\mathbf{y}) - Q_1(\mathbf{y})}{\text{RMSE}}
\end{equation}

\noindent where $\mathbf{y}$ is a column vector of observed values; $\mathbf{\hat{y}}$ is a column vector of predicted values; $N$ is the total number of observations, a scalar; $\bar{y}$ is the mean of the observed values, a scalar; $\mathbf{1}$ is a column vector of ones, with a dimension of $N \times 1$; $\| \cdot \|$ denotes the L2-norm (Euclidean norm) of a vector; $\bar{y}$ and $\bar{\hat{y}}$ are the mean of the observed and predicted values in vectors $\mathbf{y}$ and $\mathbf{\bar{y}}$, both scalars; $(\cdot)^T$ denotes the transpose of a vector; $Q_1(\mathbf{y})$ and $Q_3(\mathbf{y})$ are the first and third quartiles of the observed values in vector $\mathbf{y}$, respectively, both scalars. Please note that the sign of ME was reversed to aid in the interpretation, with a negative value indicating an average model underperformance.

To aid in the interpretation of the compressed latent space, we calculated the non-parametric Xi correlation \citep{Chatterjee2020} between each latent feature and a resampled version of the spectra. The spectra were binned every 25 cm\textsuperscript{-1} and averaged by mean to calculate the correlation. We also leveraged the Mahalanobis distance statistic ($\mathbf{d}$, Eq.~\ref{eq:d_mahalanobis}) to understand whether shifts in spectral distribution would be the cause of potential poor performance in the independent test set. $\mathbf{d}$ results can be interpreted as a multivariate generalization of the standard score, which helps understand how many standard deviations a sample is away from the training space, taking into account the covariance of the representations. After visually detecting a significant divergence from a Gaussian distribution shape in some of the latent features' probability distributions, we centered each feature at the origin and scaled to unity using the median and interquartile range.

\begin{equation}
\label{eq:d_mahalanobis}
\mathbf{d} = \sqrt{\mathbf{z}^T \mathbf{S}^{-1} \mathbf{z}}
\end{equation}
where $\mathbf{S}$ is the positive semi-definite covariance matrix of normalized latent space representations $\mathbf{z}$.

To calculate critical limits for distribution shift, a data-driven approach presented by \cite{Pomerantsev2008} and extended by \cite{Pomerantsev2013} was used in this paper. Distance values ($\mathbf{d}$) were normalized by the mean ($d_0$) and degrees of freedom ($F_d$) to leverage the Chi-Square distribution for estimating critical limits.

\begin{equation}
\label{eq:d_scaling}
d_0 = \frac{\mathbf{1}^T\mathbf{d}}{N}
\end{equation}
where $N$ is the number of observations and $\mathbf{d}$ is the Mahalanobis distance vector from the training set.

\begin{equation}
\label{eq:d_degrees}
F_d = \text{int}(\frac{2d_0^2}{\frac{1}{N}||\mathbf{d}-d_0\mathbf{1}||^2})
\end{equation}
where $F_d$ is the integer number resulting from the scaled mean and variance division.

Then, the normalized distance values will follow a Chi-square distribution with $F_d$ degrees of freedom:

\begin{equation}
\label{eq:d_normalized}
F_d\frac{\mathbf{d}}{d_0} \propto \mathcal{\chi}^2(F_d)
\end{equation}

The critical limit ($f_c$) for identifying extreme objects based on the Mahalanobis distance is computed using the inverse cumulative distribution function (ICDF), also known as the quantile function: 

\begin{equation}
\label{eq:d_normalized}
f_c = \mathcal{\chi}^2(1-\alpha, F_d)
\end{equation}
where $\alpha$ is the significance level.

We set $\alpha = 0.05$ to expect that 5\% of objects would be categorized as extreme in the training set, or when the test set feature space is compatible.


\section{Results}
\label{sec:result}
In this section, we present the  1) latent space representation; 2) provide a summary of the multi-fidelity learning and NIR to MIR conversion; 3) summarize the prediction performance from different strategies, including baselines; 4) and explore some strategies to understand potential distribution shift and the reliability of predictions.

\subsection{Latent space representation}

A 32-dimensional latent space representation of the fundamental infrared spectra (MIR) was obtained through self-supervised learning with Variational Autoencoders (VAE) (Fig.~\ref{fig:latent}, left panel). These latent features reduced the dimensionality of the spectra (D = 1,700) by approximately 50-fold, thereby decreasing redundancy and multicollinearity. The probability distribution values (PDFs) of each feature indicate that they span diverse numerical ranges and may not follow a Gaussian distribution. Some features are highly skewed and are bimodal, such as features 5, 15, 19, 24, 27, 29, and 30.

\begin{figure}[htp!]
   \centering
   \includegraphics[width=1\textwidth]{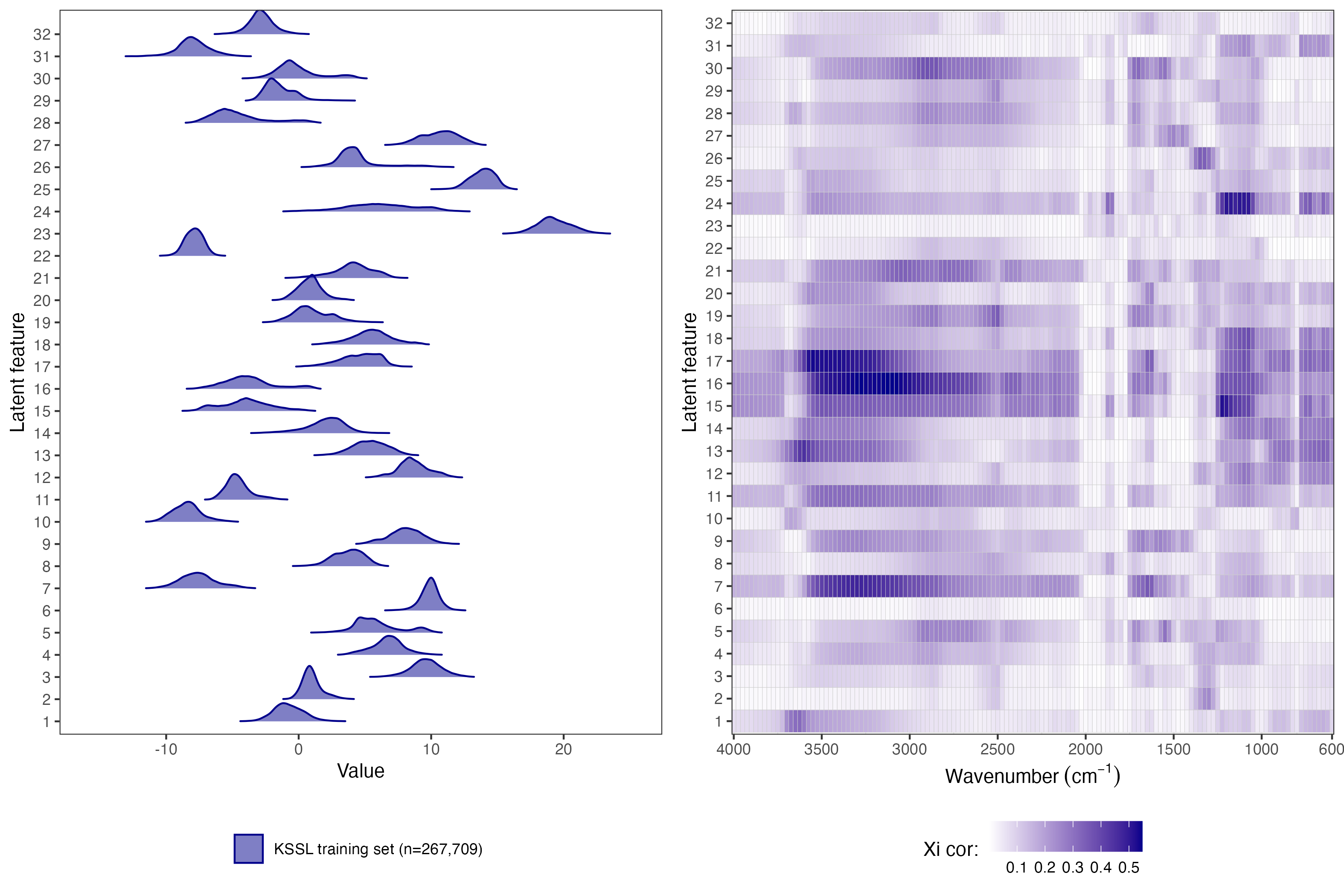}
   \caption{Probability distribution functions (PDF) of the latent space representations obtained from the self-supervised learning (left). The heatmap depicts the non-parametric Xi correlation between latent space features and the original MIR spectra, binned and averaged in increments of 25 cm\textsuperscript{-1} for the correlation analysis.}
   \label{fig:latent}
\end{figure}

The Xi correlation between latent features and MIR spectra indicates that some latent features may be partially correlated with each other, as they compressed the same signal from specific parts of the MIR spectra (e.g., features 7 and 16), while simultaneously accounting for other distinct parts of the spectra (Fig.~\ref{fig:latent}, right panel). Some latent features represent broader parts of the spectra (e.g., features 7, 11, 15, 16, 17, 21, 24, 30, etc.), while others were able to emphasize the variability of specific spectral features (e.g., features 1, 2, 6, 10, 19, 22, 23, etc.). For example, the variation captured by latent feature two, which initially maps to spectral variations around 1400, 2500, and 2800 cm\textsuperscript{-1}, is associated with soil samples containing high concentrations of carbonates. The fundamental vibration of the CO\textsubscript{3}\textsuperscript{2-} group is centered around 1450 cm\textsuperscript{-1} with weaker overtones and combination vibrations around 2500 and 2800 cm\textsuperscript{-1} \citep{Farmer1975}.

\subsection{Multi-fidelity learning}

\begin{figure}[H]
   \centering
   \includegraphics[width=1\textwidth]{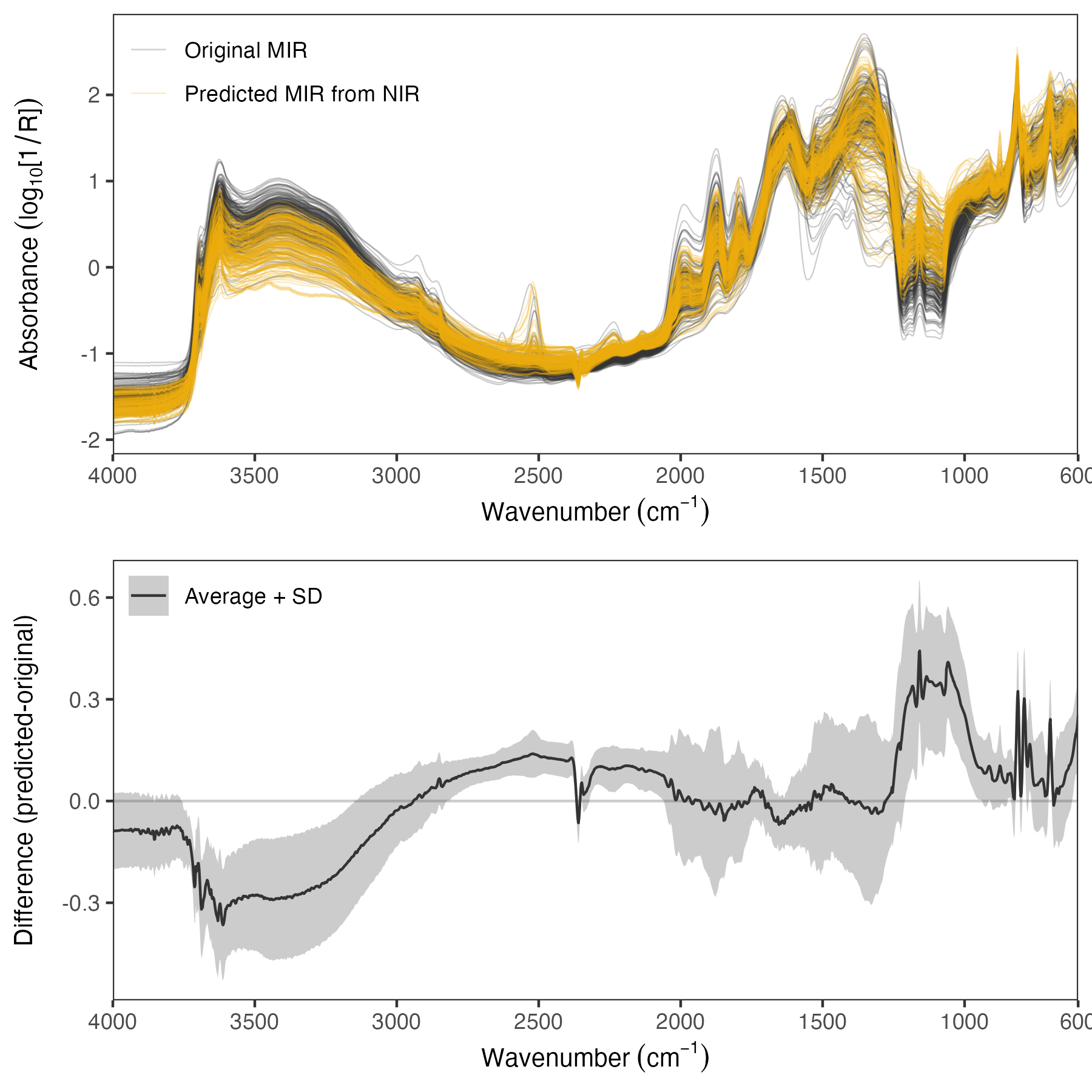}
   \caption{Comparison of original and predicted mid-infrared (MIR) spectra of the independent test samples (top panel, n=206). Mean and standard deviation difference (predicted-observed) is highlighted in the bottom panel.}
   \label{fig:conversion}
\end{figure}

The second task of this study was to bridge the two infrared spectra ranges (NIR and MIR) through a unified latent space representation. No apparent major shift is observed when we visualize the original and predicted MIR (from NIR) for the independent test samples (Fig.~\ref{fig:conversion}, top panel). Most of the spectral features are preserved during the conversion, although some offsets at specific parts of the spectra are more pronounced. Despite the good overall correspondence between the original and predicted versions, the dissimilarity of specific locations becomes evident when we plot the difference between both spectra across the MIR range (Fig.~\ref{fig:conversion}, bottom panel). The regions between 3750-3000 cm\textsuperscript{-1} and around 1200 cm\textsuperscript{-1} are affected by a systematic deviation from the expected spectral variation, with the first region being underpredicted while the second being overpredicted. Similarly, the features around 2000 and 1250 cm\textsuperscript{-1} exhibit significant variance compared to the other spectral features, although without an apparent bias, as previously indicated. We further investigate whether and to what degree these apparent shifts and significant variance impact prediction performance across a range of soil properties.

\subsection{Predictive soil spectroscopy}

\begin{figure}[H]
   \centering
   \includegraphics[width=1\textwidth]{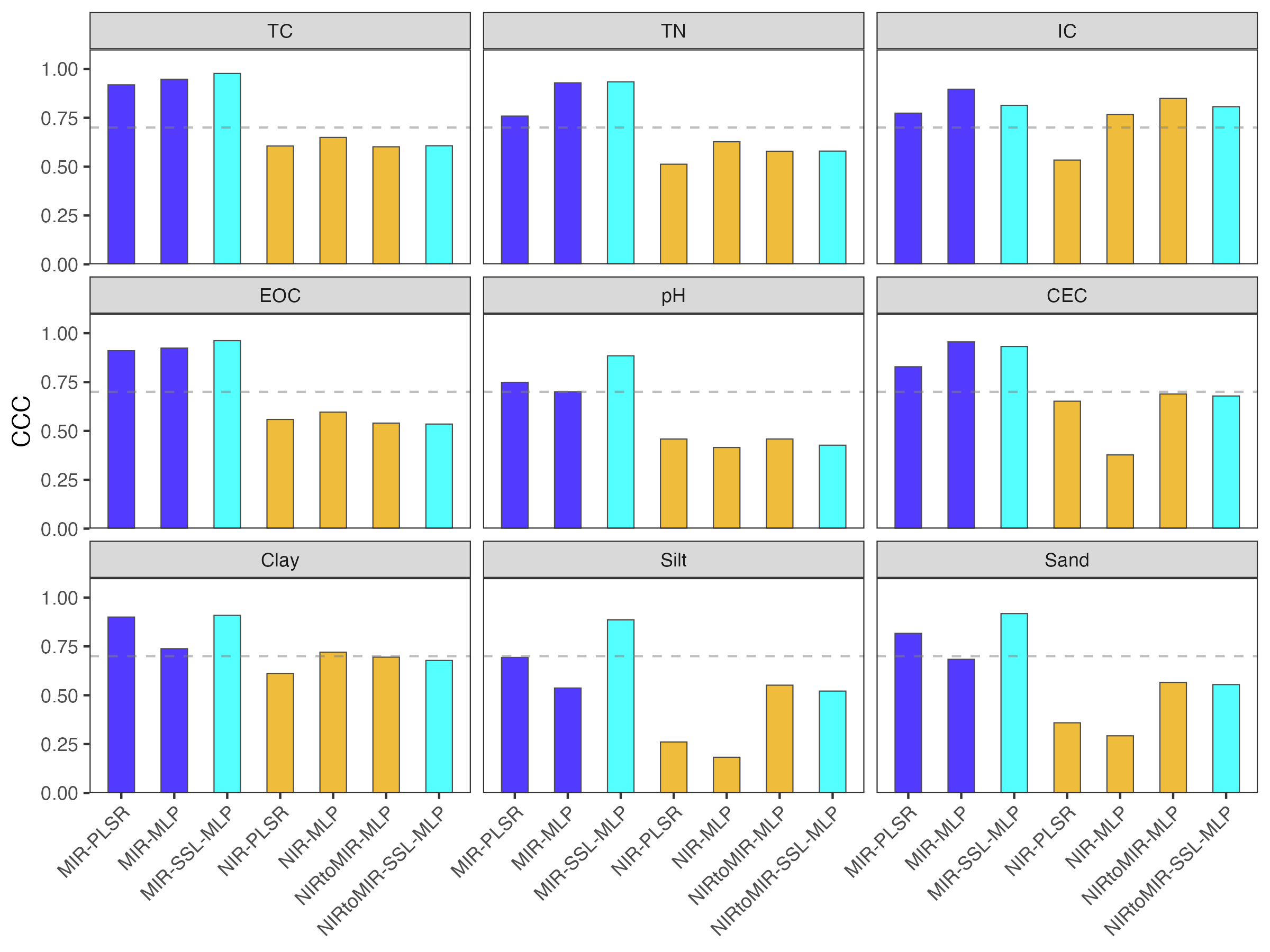}
   \caption{Lin's concordance correlation coefficient (CCC) of different strategies for soil properties prediction of an independent test set. Reference models are provided for both near-infrared (NIR) and mid-infrared (MIR) spectra using two fitting algorithms: partial least squares regression (PLSR) and multilayer perceptron (MLP). In addition, MLP is combined with latent space embeddings derived from the self-supervised learning task (SSL) of MIR spectra, as well as with both the predicted MIR spectra and their latent space embeddings derived from multi-fidelity learning (NIRtoMIR). A horizontal line is fixed at CCC = 0.7. Blue color represents prediction models using MIR spectrum as input. Cyan color represents prediction models using latent space representations as input. And Yellow colors  represents prediction models using NIR spectrum as inputs.}
   \label{fig:ccc}
\end{figure}

\begin{figure}[H]
   \centering
   \includegraphics[width=1\textwidth]{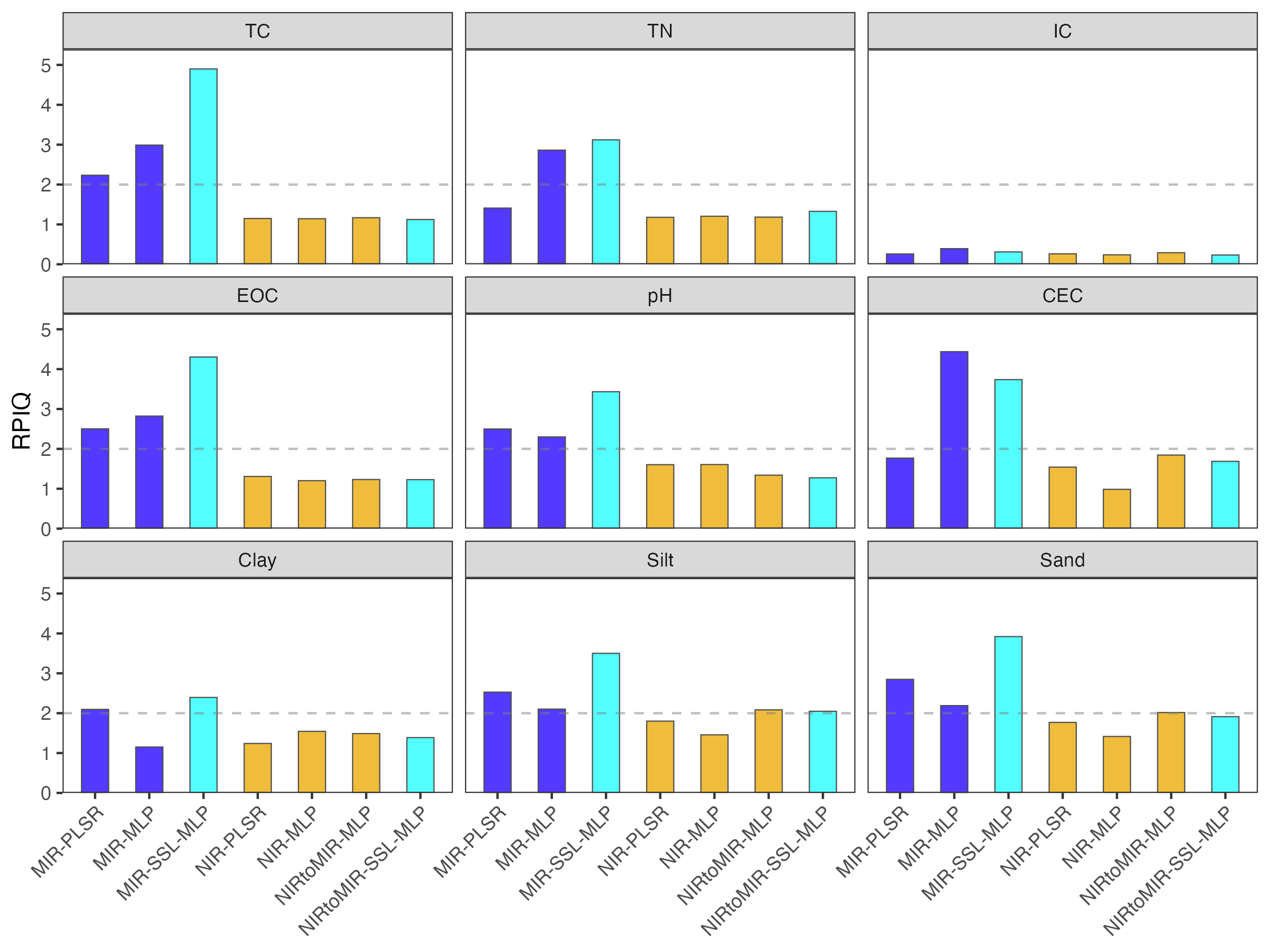}
   \caption{Ratio of performance to the interquartile range (RPIQ) of different strategies for soil properties prediction of an independent test set. Reference models are provided for both near-infrared (NIR) and mid-infrared (MIR) spectra using two fitting algorithms: partial least squares regression (PLSR) and multilayer perceptron (MLP). In addition, MLP is combined with latent space embeddings derived from the self-supervised learning task (SSL) of MIR spectra, as well as with both the predicted MIR spectra and their latent space embeddings derived from multi-fidelity learning (NIRtoMIR). A horizontal line is fixed at RPIQ = 2. Blue: Prediction based on MIR spectrum. Cyan: Prediction based on latent space representation. Yellow: Prediction based on NIR spectrum.}
   \label{fig:rpiq}
\end{figure}

Model evaluation on an independent test set, which comprises quality control samples representing cropland soils across the USA, revealed that MIR-based models consistently yielded superior results than NIR-based models (Fig.~\ref{fig:ccc}). Baseline MIR models achieved similar Lin's CCC values regardless of the fitting algorithm, with an average CCC of 0.82 for PLSR and 0.81 for MLP across nine soil properties. MLP was superior for TC (CCC of 0.94 vs 0.92), TN (CCC of 0.93 vs 0.76), IC (CCC of 0.90 vs 0.77), EOC (CCC of 0.92 vs 0.91), and CEC (CCC of 0.96 vs 0.83), while PLSR resulted in better models for pH (CCC of 0.75 vs 0.70), Clay (CCC of 0.90 vs 0.83), Silt (CCC of 0.70 vs 0.54), and Sand (CCC of 0.82 vs 0.68). This analysis also revealed that after learning latent representations from a large and diverse MIR spectral library (KSSL MIR dataset) via self-supervised learning, an MLP prediction model trained with the resulting latent space embeddings ($D=32$, MIR-SSL-MLP) achieved a higher CCC (average 0.91) across all soil properties, except for IC and CEC, with a CCC slightly lower than MIR-MLP. For MIR-SSL-MLP, only Silt (CCC of 0.87), pH (CCC of 0.88), and IC (CCC of 0.81) did not surpass an accuracy threshold of higher than 0.9 CCC. The superiority of MIR-SSL-MLP also became evident when comparing RPIQ across all modeling strategies (Fig.~\ref{fig:rpiq}), with MIR-SSL-MLP consistently outperforming the two baseline models (MIR-PLSR and MIR-MLP) except for CEC and IC.

NIR-based models, in turn, yielded average CCC between 0.51 and 0.61 depending on the prediction strategy. Baseline models with PLSR and MLP achieved an average CCC of 0.51, whereas the multifidelity learning strategies yielded CCCs of 0.61 and 0.60 for NIRtoMIR-MLP and NIRtoMIR-SSL-MLP, respectively. Performance enhancement via multi-fidelity learning was found for IC (CCC of 0.85 with NIRtoMIR-MLP, as compared to the best baseline of CCC = 0.77 with NIR-MLP), CEC (CCC of 0.69 with NIRtoMIR-MLP, as compared to the best baseline of CCC = 0.65 with NIR-PLSR), Silt (CCC of 0.55 with NIRtoMIR-MLP, as compared to the best baseline of CCC = 0.26 with NIR-MLP), and Sand (CCC of 0.57 with NIRtoMIR-MLP, as compared to the best baseline of CCC = 0.36 with NIR-MLP). Nevertheless, although the multifidelity learning strategies (NIRtoMIR) were not favorable for predicting other soil properties, they achieved almost the same performance in terms of CCC as one or both of the baselines.

It is worth noting, however, that based on the RPIQ metric, the prediction performance results may be less favorable than those evaluated by CCC. RPIQ is a standardized metric for the RMSE that enables an intercomparison between different modeling strategies and soil properties, as it is not affected by the measurement unit and range of the outcomes. A value of 1 indicates that the original variability of the soil property, as quantified by IQR (50\% of the probability distribution), is the same as the average model prediction error. A threshold of 2 indicates that the average model error is at least half of the variability present in the soil property, suggesting that the models are sensitive enough to differentiate the variability in the dataset. When averaging each modeling strategy across the nine soil properties, all models achieved an RPIQ of at least 1, with the NIR baselines and NIRtoMIR-based models ranging between 1.20 (NIR-MLP) and 1.41 (NIRtoMIR-MLP), and models based on MIR ranging between 2.02 (MIR-PLSR) and 3.29 (MIR-SSL-MLP). This indicates that, although all models achieved intermediate to satisfactory levels for CCC and an RPIQ of at least 1, NIR-based models may struggle in soil prediction tasks due to their lower spectral sensitivity (low fidelity) and prediction consistency compared to MIR. Regardless of the modeling strategy and spectral range, IC performed poorly when evaluated by RPIQ, and potential causes are discussed further in the paper. Lastly, all performance metrics for the independent test set are provided in \ref{app_performance}.

\subsection{Reliability of predictions}

\begin{figure}[H]
   \centering
   \includegraphics[width=1\textwidth]{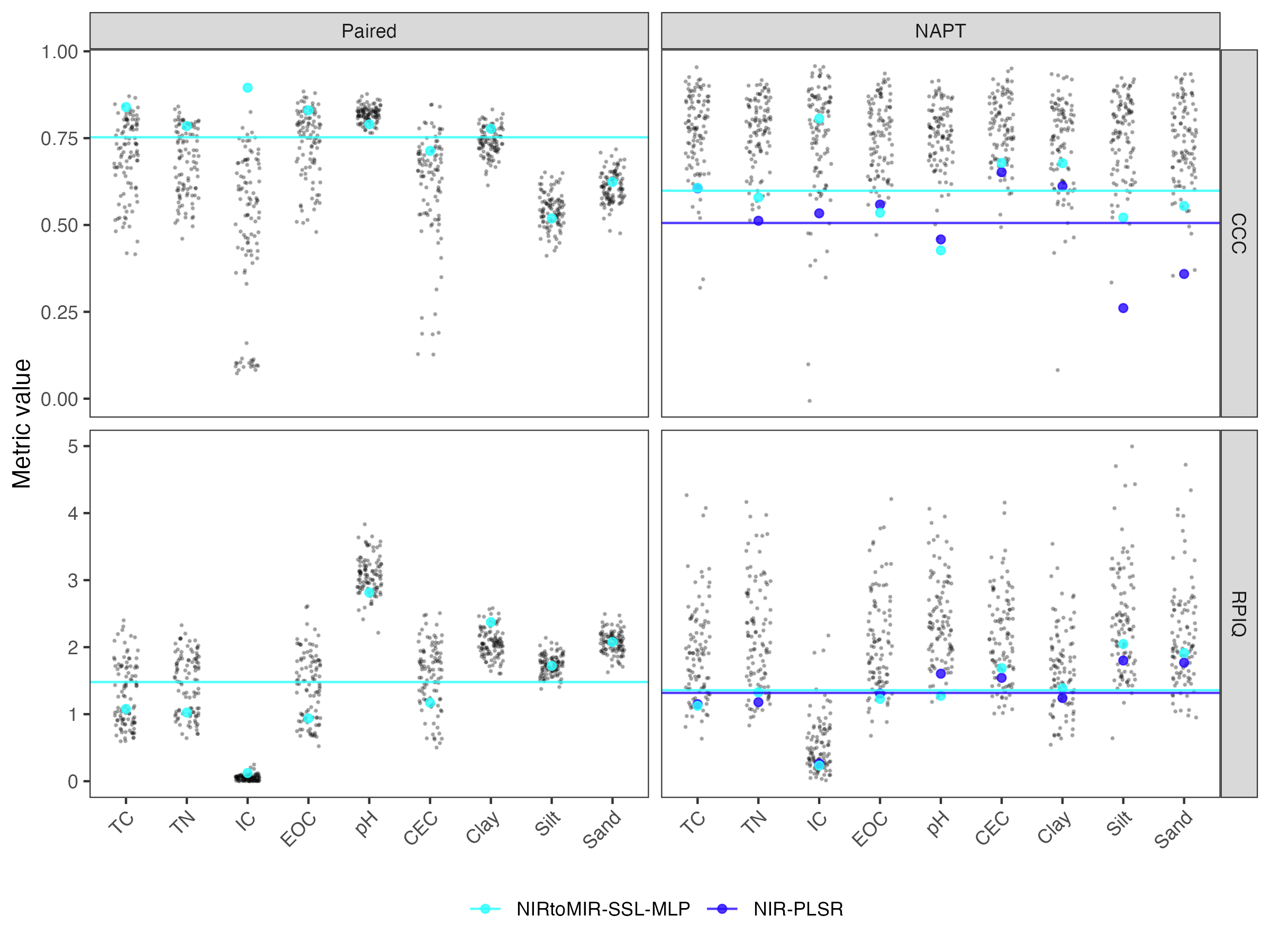}
   \caption{Paired (validation split) and NAPT (independent test set) performance -- Lin's concordance correlation coefficient (CCC) and ratio of performance to the interquartile range (RPIQ) -- for a baseline (NIR-PLSR) and advanced modeling strategies (NItoMIR-SSL-MLP). Validation and test performance are overlaid on top of each dataset's internal performance estimated by 10-fold cross-validation repeated 10 times with partial least squares regression (PLSR).}
   \label{fig:eval_perf_shift}
\end{figure}

The systematic lower performance of NIR-based models was further investigated with an additional analysis (Fig.~\ref{fig:eval_perf_shift}). We compared the performance of our prediction models on two different datasets: the random validation split (used during model development, as described in Paired) and the independent NAPT test set. Our analysis revealed that models performed better on the Paired dataset validation split than on the NAPT dataset, regardless of the prediction method employed (NIR-to-MIR or NIR-PLSR strategy). In fact, a slight performance edge was obtained when using the converted spectra (NIRtoMIR) as compared to using the original NIR, possibly because the MIR-based model was trained with a larger and more diverse soil dataset. When we evaluated the internal performance of each dataset using 10-fold cross-validation repeated 10 times (background dots), we found that the NIR-based models exhibited a wide range of performance, from poor to satisfactory, with NAPT showing higher adherence when trained internally (dots more concentrated on upper levels for both CCC and RPIQ). While the performance on the Paired validation split fell within this expected range, the independent performance on the NAPT dataset was consistently lower than its internal performance range. This systematic decrease suggests a potential mismatch between the NAPT and Paired datasets, possibly due to differences in either the spectral representation or the range of soil properties.

\subsection{Ablation Study: Isometric Log Ratio Transformation}
\begin{figure}[H]
\centering
\includegraphics[width=0.79\textwidth]{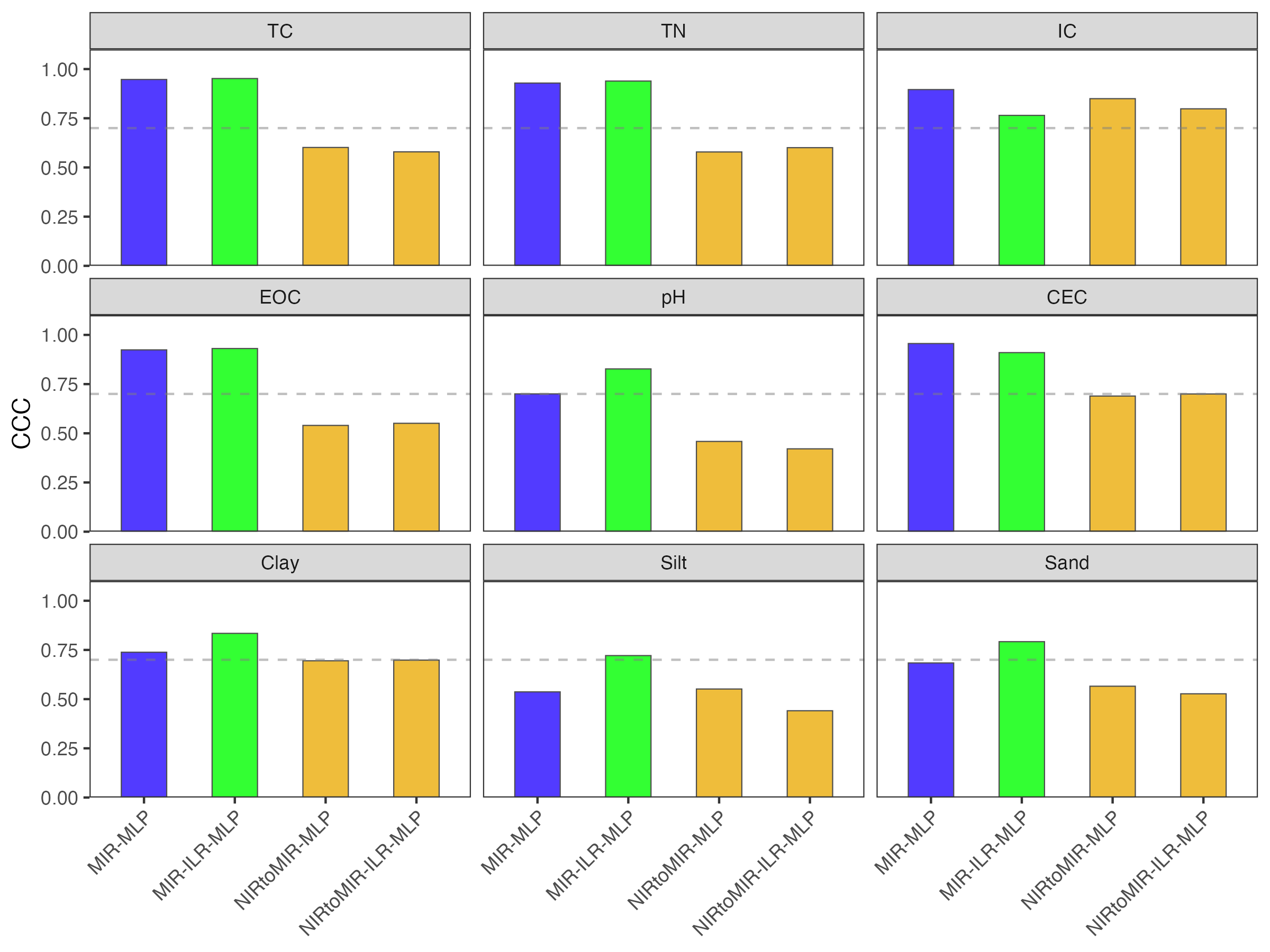}
\caption{Ablation study, the effect of isometric log ratio (ILR). Green color represents model training with ILR transformation.}
\label{fig:ilr_R2}
\end{figure}

We also demonstrated the effect of knowledge-informed learning in the current dataset. Specifically, we trained another multi-task learning NN surrogate to predict soil properties based on MIR spectrum. During training, we reduced the three soil texture variables (i.e., clay, sand, and silt) to two using an isometric log ratio transformation based on Eq.~\ref{eq:ilr}. To make inferences, we then used the inverse isometric log ratio to restore the the three soil particle size distribution variables. This transformation strictly enforced the constraint that the sum of clay, silt and sand is always equal to $100\%$. The comparison result is shown in Fig.~\ref{fig:ilr_R2}. It shows that including the soil components constraints (MLR-ILR-MLP) improves the prediction result of clay, silt and sand components in baseline model (MLR-MLP). Furthermore, we also evaluated the more challenging task of predicting MIR first then predicting soil properties (model NIRtoMIR-MLP and NIRtoMIR-ILR-MLP). However, we found that including this ILR transformation actually worsened the result on the independent validation dataset. The reason for this reduced performance is that the bottleneck is still the spectrum conversion. Moreover, additional ablation study on NN architecture is listed in Sec.~\ref{sec:NN_ablation}.

\section{Discussion}
\label{sec:discussion}

This work shows that a self-supervised \emph{representation learning} framework trained on a large MIR dataset can learn a compact latent space that benefits soil spectroscopy in two primary ways. First, it improves prediction accuracy and consistency across diverse soil properties. Second, it provides a bridge that maps lower-fidelity NIR measurements into a higher-fidelity, MIR-calibrated representation via a shared latent manifold. By leveraging unlabeled MIR scan repeats during pretraining, we compress the spectral representation from $\sim$1{,}700 channels to a 32-dimensional latent space while preserving chemically meaningful structure, yielding gains over standard pre-processing methods. Moreover, models trained primarily on MIR transfer to NIR through the shared latent space without degradation relative to NIR-only baselines; in several properties the transferred model outperforms the NIR-MLP baseline.

The value of the proposed framework is threefold. First, performance gains. On the independent test set, the MIR-embeddings + MLP model (MIR-SSL-MLP) achieves the highest scores across nine soil properties, indicating that the learned latent space retains distinctive, property-relevant features while suppressing redundancy. Second, data efficiency. Because pretraining is self-supervised, repeated MIR scans serve as effective augmentation, reducing labeling demand for downstream calibration and enabling reuse of a single pretrained latent space across properties and sites. Third, interpretability. We interpret the MIR-trained latent with Xi band correlations (Fig.~\ref{fig:latent}) and map NIR data into the same latent manifold, allowing NIR predictions to inherit MIR-grounded feature interpretations.

A key observation for NIR--MIR conversion is that transferring MIR-trained models to NIR via the shared latent space does not degrade performance relative to NIR-only baselines and, for several properties, improves it. This suggests that the learned latent space captures the fundamental linkage strongly enough that NIR behaves \emph{as if} it carried MIR information for downstream prediction. For portable, lower-cost NIR deployments, especially when only limited labels and small paired NIR--MIR subsets are available, this provides a practical path to leverage the much larger and diverse MIR datasets without paying an accuracy penalty compared with standard NIR workflows.


\section*{Acknowledgements}
This work was performed under the auspices of the U.S. Department of Energy by the Lawrence Livermore National Laboratory under Contract No. DE-AC52-07NA27344. We acknowledge support from
LLNL Laboratory Directed Research and Development (LDRD) Program Grant No. 24-SI-002. This project is also partially supported by DOE ECRP (51917/SCW1885). The work is reviewed and
released under LLNL-JRNL-2012899.
\newpage
\appendix
\section{Table: Summary statistics of nine soil properties across three different datasets.}
\label{app_soil}
\begin{longtable}{llrrrrrrr}
\hline
  \hline
Property & Dataset & n & min & mean & sd & max & skewness & kurtosis \\ 
  \hline
TC & KSSL & 51508 & 0.00 & 2.36 & 4.48 & 62.43 & 6.40 & 57.73 \\ 
  TC & Paired & 2016 & 0.02 & 2.29 & 2.80 & 53.88 & 8.25 & 124.40 \\ 
  TC & NAPT & 206 & 0.25 & 1.88 & 1.25 & 7.71 & 1.53 & 6.08 \\ 
  TN & KSSL & 51508 & 0.00 & 0.16 & 0.35 & 41.90 & 44.20 & 4604.89 \\ 
  TN & Paired & 2016 & 0.00 & 0.17 & 0.18 & 3.02 & 6.53 & 84.59 \\ 
  TN & NAPT & 206 & 0.03 & 0.16 & 0.09 & 0.66 & 1.58 & 7.68 \\ 
  IC & KSSL & 51508 & 0.00 & 3.34 & 9.47 & 105.77 & 4.49 & 28.86 \\ 
  IC & Paired & 2016 & 0.00 & 2.03 & 6.23 & 89.03 & 5.16 & 41.93 \\ 
  IC & NAPT & 206 & 0.20 & 1.79 & 3.27 & 22.41 & 3.48 & 16.46 \\ 
  EOC & KSSL & 51508 & 0.00 & 1.95 & 4.41 & 62.43 & 6.88 & 63.82 \\ 
  EOC & Paired & 2016 & 0.00 & 2.05 & 2.75 & 53.88 & 8.79 & 135.82 \\ 
  EOC & NAPT & 206 & 0.28 & 1.73 & 1.12 & 7.53 & 1.69 & 7.76 \\ 
  pH & KSSL & 51508 & 2.29 & 6.53 & 1.31 & 10.70 & 0.03 & 2.10 \\ 
  pH & Paired & 2016 & 3.69 & 6.28 & 1.26 & 9.52 & 0.14 & 1.91 \\ 
  pH & NAPT & 206 & 4.82 & 6.77 & 0.88 & 8.30 & 0.00 & 2.03 \\ 
  CEC & KSSL & 51508 & 0.00 & 17.22 & 16.18 & 584.59 & 5.23 & 73.48 \\ 
  CEC & Paired & 2016 & 0.13 & 16.23 & 12.12 & 190.11 & 4.03 & 44.85 \\ 
  CEC & NAPT & 206 & 2.30 & 16.40 & 7.99 & 44.70 & 0.71 & 3.44 \\ 
  Clay & KSSL & 51508 & 0.00 & 22.64 & 16.07 & 96.14 & 0.81 & 3.43 \\ 
  Clay & Paired & 2016 & 0.00 & 19.86 & 13.91 & 86.69 & 0.88 & 3.83 \\ 
  Clay & NAPT & 206 & 1.97 & 18.37 & 10.23 & 54.11 & 1.18 & 4.92 \\ 
  Silt & KSSL & 51508 & 0.00 & 38.10 & 20.38 & 94.50 & 0.01 & 2.17 \\ 
  Silt & Paired & 2016 & 0.00 & 38.32 & 20.26 & 87.90 & 0.16 & 2.29 \\ 
  Silt & NAPT & 206 & 1.59 & 44.07 & 21.10 & 79.00 & 0.00 & 1.93 \\ 
  Sand & KSSL & 51508 & 0.10 & 39.26 & 29.20 & 100.00 & 0.47 & 2.00 \\ 
  Sand & Paired & 2016 & 0.30 & 41.82 & 28.19 & 100.00 & 0.28 & 1.94 \\ 
  Sand & NAPT & 206 & 3.00 & 37.56 & 26.16 & 96.39 & 0.62 & 2.16 \\ 
   \hline
\end{longtable}
\noindent Soil properties: Total carbon (TC), total nitrogen (TN), inorganic carbon (IC), estimated organic carbon (EOC), pH, cation exchange capacity (CEC), clay, silt, and sand. Datasets: The Kellogg Soil Survey Laboratory mid-infrared (MIR) database (KSSL), a subset of the KSSL database containing both near-infrared (NIR) and MIR spectra (Paired), and the soil archive from the North American Proficiency Testing (NAPT) program, which was scanned for both MIR and NIR.

\section{Table: Prediction performance metrics on the independent test set with different modeling strategies and across nine soil properties.} 
\label{app_performance}
\begin{longtable}{llrrrrr}
  \hline
Property & Strategy & RMSE & ME & R\textsuperscript{2} & CCC & RPIQ \\ 
  \hline
TC & MIR-PLSR & 0.58 & 0.32 & 0.94 & 0.92 & 2.23 \\ 
  TC & MIR-MLP & 0.43 & 0.30 & 0.96 & 0.95 & 2.99 \\ 
  TC & MIR-SSL-MLP & 0.26 & -0.01 & 0.95 & 0.98 & 4.90 \\ 
  TC & NIR-PLSR & 1.13 & 0.52 & 0.44 & 0.61 & 1.15 \\ 
  TC & NIR-MLP & 1.13 & -0.41 & 0.47 & 0.65 & 1.14 \\ 
  TC & NIRtoMIR-MLP & 1.11 & 0.45 & 0.42 & 0.60 & 1.17 \\ 
  TC & NIRtoMIR-SSL-MLP & 1.15 & 0.47 & 0.42 & 0.61 & 1.12 \\ 
  TN & MIR-PLSR & 0.08 & 0.06 & 0.85 & 0.76 & 1.41 \\ 
  TN & MIR-MLP & 0.04 & 0.02 & 0.90 & 0.93 & 2.86 \\ 
  TN & MIR-SSL-MLP & 0.03 & -0.02 & 0.91 & 0.93 & 3.12 \\ 
  TN & NIR-PLSR & 0.09 & 0.06 & 0.44 & 0.51 & 1.18 \\ 
  TN & NIR-MLP & 0.09 & -0.04 & 0.46 & 0.63 & 1.21 \\ 
  TN & NIRtoMIR-MLP & 0.09 & 0.05 & 0.44 & 0.58 & 1.18 \\ 
  TN & NIRtoMIR-SSL-MLP & 0.08 & 0.04 & 0.43 & 0.58 & 1.33 \\ 
  IC & MIR-PLSR & 2.49 & -0.59 & 0.64 & 0.77 & 0.26 \\ 
  IC & MIR-MLP & 1.65 & -0.51 & 0.84 & 0.90 & 0.39 \\ 
  IC & MIR-SSL-MLP & 2.08 & -0.52 & 0.68 & 0.81 & 0.31 \\ 
  IC & NIR-PLSR & 2.46 & -0.59 & 0.57 & 0.53 & 0.26 \\ 
  IC & NIR-MLP & 2.72 & -0.14 & 0.66 & 0.77 & 0.24 \\ 
  IC & NIRtoMIR-MLP & 2.23 & -0.17 & 0.83 & 0.85 & 0.29 \\ 
  IC & NIRtoMIR-SSL-MLP & 2.78 & 0.25 & 0.83 & 0.81 & 0.23 \\ 
  EOC & MIR-PLSR & 0.53 & 0.32 & 0.92 & 0.91 & 2.50 \\ 
  EOC & MIR-MLP & 0.47 & 0.29 & 0.92 & 0.92 & 2.82 \\ 
  EOC & MIR-SSL-MLP & 0.31 & 0.09 & 0.93 & 0.96 & 4.30 \\ 
  EOC & NIR-PLSR & 1.02 & 0.46 & 0.38 & 0.56 & 1.31 \\ 
  EOC & NIR-MLP & 1.11 & -0.43 & 0.41 & 0.60 & 1.20 \\ 
  EOC & NIRtoMIR-MLP & 1.08 & 0.39 & 0.33 & 0.54 & 1.23 \\ 
  EOC & NIRtoMIR-SSL-MLP & 1.09 & 0.44 & 0.34 & 0.54 & 1.23 \\ 
  pH & MIR-PLSR & 0.58 & 0.03 & 0.58 & 0.75 & 2.50 \\ 
  pH & MIR-MLP & 0.63 & -0.18 & 0.54 & 0.70 & 2.30 \\ 
  pH & MIR-SSL-MLP & 0.42 & -0.10 & 0.79 & 0.88 & 3.43 \\ 
  pH & NIR-PLSR & 0.90 & -0.32 & 0.24 & 0.46 & 1.60 \\ 
  pH & NIR-MLP & 0.90 & -0.39 & 0.23 & 0.42 & 1.61 \\ 
  pH & NIRtoMIR-MLP & 1.07 & -0.60 & 0.31 & 0.46 & 1.34 \\ 
  pH & NIRtoMIR-SSL-MLP & 1.13 & -0.64 & 0.28 & 0.43 & 1.27 \\ 
  CEC & MIR-PLSR & 6.02 & 3.72 & 0.90 & 0.83 & 1.77 \\ 
  CEC & MIR-MLP & 2.40 & 1.12 & 0.93 & 0.96 & 4.44 \\ 
  CEC & MIR-SSL-MLP & 2.85 & -0.27 & 0.87 & 0.93 & 3.74 \\ 
  CEC & NIR-PLSR & 6.90 & -1.67 & 0.44 & 0.65 & 1.54 \\ 
  CEC & NIR-MLP & 10.80 & -8.97 & 0.45 & 0.38 & 0.99 \\ 
  CEC & NIRtoMIR-MLP & 5.77 & -0.49 & 0.49 & 0.69 & 1.84 \\ 
  CEC & NIRtoMIR-SSL-MLP & 6.31 & 0.61 & 0.46 & 0.68 & 1.69 \\ 
  Clay & MIR-PLSR & 5.09 & 0.73 & 0.85 & 0.90 & 2.09 \\ 
  Clay & MIR-MLP & 9.25 & 7.09 & 0.81 & 0.74 & 1.15 \\ 
  Clay & MIR-SSL-MLP & 4.45 & 1.25 & 0.84 & 0.91 & 2.39 \\ 
  Clay & NIR-PLSR & 8.59 & 4.92 & 0.53 & 0.61 & 1.24 \\ 
  Clay & NIR-MLP & 6.90 & -2.83 & 0.62 & 0.72 & 1.54 \\ 
  Clay & NIRtoMIR-MLP & 7.16 & 2.33 & 0.56 & 0.69 & 1.49 \\ 
  Clay & NIRtoMIR-SSL-MLP & 7.69 & 2.94 & 0.53 & 0.68 & 1.39 \\ 
  Silt & MIR-PLSR & 13.41 & -0.38 & 0.65 & 0.69 & 2.53 \\ 
  Silt & MIR-MLP & 16.13 & -1.99 & 0.45 & 0.54 & 2.10 \\ 
  Silt & MIR-SSL-MLP & 9.68 & -4.20 & 0.83 & 0.89 & 3.50 \\ 
  Silt & NIR-PLSR & 18.83 & -0.66 & 0.25 & 0.26 & 1.80 \\ 
  Silt & NIR-MLP & 23.26 & -10.86 & 0.08 & 0.18 & 1.46 \\ 
  Silt & NIRtoMIR-MLP & 16.27 & 5.18 & 0.50 & 0.55 & 2.08 \\ 
  Silt & NIRtoMIR-SSL-MLP & 16.55 & 2.24 & 0.40 & 0.52 & 2.05 \\ 
  Sand & MIR-PLSR & 14.23 & -2.07 & 0.71 & 0.82 & 2.85 \\ 
  Sand & MIR-MLP & 18.51 & -6.72 & 0.56 & 0.68 & 2.19 \\ 
  Sand & MIR-SSL-MLP & 10.34 & 3.16 & 0.86 & 0.92 & 3.92 \\ 
  Sand & NIR-PLSR & 22.92 & -4.09 & 0.26 & 0.36 & 1.77 \\ 
  Sand & NIR-MLP & 28.62 & 15.30 & 0.17 & 0.29 & 1.42 \\ 
  Sand & NIRtoMIR-MLP & 20.11 & -6.28 & 0.48 & 0.57 & 2.02 \\ 
  Sand & NIRtoMIR-SSL-MLP & 21.19 & -6.25 & 0.40 & 0.55 & 1.91 \\ 
   \hline
\end{longtable}
\noindent Soil properties: Total carbon (TC), total nitrogen (TN), inorganic carbon (IC), estimated organic carbon (EOC), pH, cation exchange capacity (CEC), clay, silt, and sand. Modeling strategies: mid-infrared (MIR) spectra mapped directly to soil properties with partial least squares regression (MIR-PLSR) or multilayer perceptron (MIR-MLP); MIR latent space embedding mapped to soil properties with MLP (MIR-SSL-MLP); near-infrared (NIR) spectra mapped directly to soil properties with partial least squares regression (NIR-PLSR) or multilayer perceptron (NIR-MLP); NIR mapped to MIR from multi-fidelity learning, then mapped to soil properties with predicted MIR (NIRtoMIR-MLP) or latent space embeddings (NIRtoMIR-SSL-MLP).

\section{Ablation study: neural network architecture}
\label{sec:NN_ablation}
\begin{figure}[H]
\centering
\includegraphics[width=0.75\textwidth]{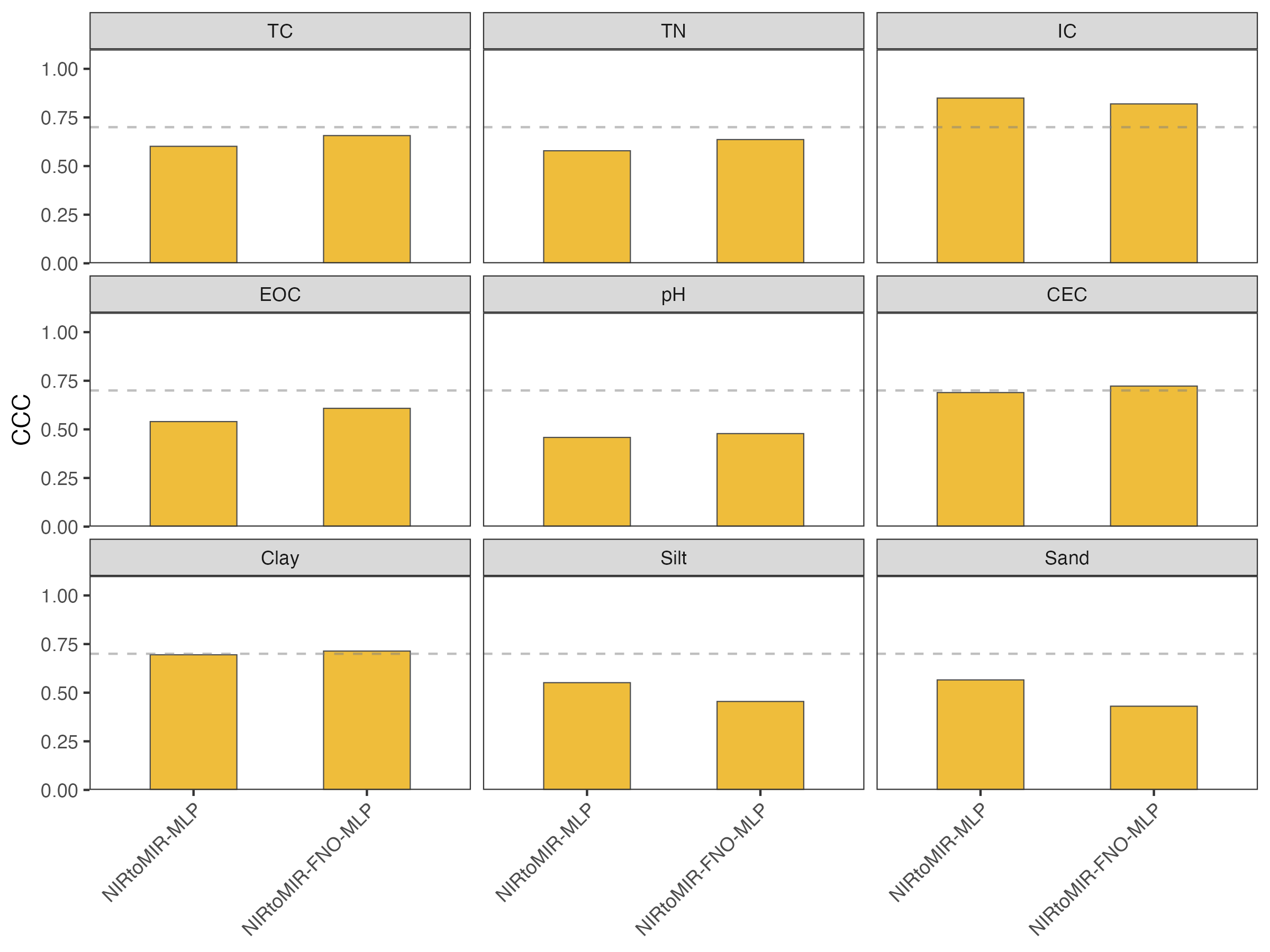}
\caption{NN architecture comparison}
\label{fig:FNO_R2}
\end{figure}

We further tested if various NN architecture could improve the spectrum conversion between NIR to MIR. Recently, the FNO framework~\cite{li2020fourier} gained increasing popularity by demonstrating that it could improve the learning result in the frequency space. We replaced the MLP model from NIR to MIR with a FNO architecture while keeping the same surrogate model from MIR spectrum to soil properties. The FNO architecture is not compatible with the framework of latent space, so we trained it from scratch. The comparison result in Fig.~\ref{fig:FNO_R2} shows that the FNO performance in all the predicted models is inferior to the MLP architecture. It is understandable that based on Occam's razor algorithms (also known as the law of parsimony), the increased complexity in models does not guarantee a better explanation of the data. This comparison also justifies why we did not use more sophisticated architectures (e.g., FNO, transformer) for the current tasks.
\newpage
\bibliographystyle{elsarticle-harv} 
\bibliography{ref}
\end{document}